# Knowledge Combination in Graphical Multiagent Models

**Quang Duong** **Michael P. Wellman** **Satinder Singh**

University of Michigan
Computer Science & Engineering
Ann Arbor, MI 48109-2121 USA
{qduong,wellman,baveja}@umich.edu

## Abstract

A *graphical multiagent model* (GMM) represents a joint distribution over the behavior of a set of agents. One source of knowledge about agents' behavior may come from game-theoretic analysis, as captured by several graphical game representations developed in recent years. GMMs generalize this approach to express arbitrary distributions, based on game descriptions or other sources of knowledge bearing on beliefs about agent behavior. To illustrate the flexibility of GMMs, we exhibit game-derived models that allow probabilistic deviation from equilibrium, as well as models based on heuristic action choice. We investigate three different methods of integrating these models into a single model representing the combined knowledge sources. To evaluate the predictive performance of the combined model, we treat as actual outcome the behavior produced by a reinforcement learning process. We find that combining the two knowledge sources, using any of the methods, provides better predictions than either source alone. Among the combination methods, mixing data outperforms the opinion pool and direct update methods investigated in this empirical trial.

## 1 INTRODUCTION

Graphical models provide a compact representation for domains with decomposable structure, with concomitant computational advantages. Multiagent scenarios may be particularly amenable to decomposition, to the extent that interactions among the agents exhibit localized effects. The idea of exploiting conditional independence among the effects of agents' decisions was central to the multiagent influence diagram (MAID) framework developed by Koller and Milch (2003). This observation was also a driving motivation for *graphical game* models, first introduced by Kearns et al. (2001), and subsequently examined and extended in several research efforts (Kearns, 2007).

In the basic graphical game approach, the model is a factored representation of a normal-form game, and special-purpose algorithms operate on this representation to identify approximate or exact Nash equilibria. Daskalakis and Papadimitriou (2006) demonstrated how to map a graphical game to a Markov random field (MRF), assigning high potential to configurations where an agent plays a best response to its neighbors. They showed that the maximum a posteriori configurations of the MRF correspond to pure-strategy Nash equilibria (PSNE) of the game. This approach enables the exploitation of statistical inference tools for game-theoretic computation, including the full repertoire of graphical model algorithms.

We build on these works to introduce *graphical multiagent models* (GMMs), which are simply graphical models where the joint probability distribution is interpreted as an uncertain belief (e.g., a prediction) about the agents' play. For instance, the Daskalakis and Papadimitriou mapping can be viewed as a GMM where we believe that the agents will play a PSNE if one exists. This is of course just one candidate for belief based on game-theoretic analysis. When reasoning about strategies to play, or designing a mechanism (which induces a game for other agents), we may wish to adopt alternative bases for forming beliefs about the agents' play (Vorobeychik and Wellman, 2006). The GMM framework supports such decision making, and moreover, allows that beliefs may be based on variant solution concepts, models of bounded rationality or equilibrium selection, or for that matter knowledge that has nothing to do with game-theoretic analysis. At this level, our motivation shares the spirit of the network of influence diagrams formalism of Gal and Pfeffer (2008), which extends MAIDs to incorporate non-maximizing models of behavior. It also aligns with the goal of Wolpert's information-theoretic framework for modeling bounded rationality in game play (Wolpert, 2006).

In this paper, we illustrate the flexibility of GMMs by showing how to construct plausible multiagent models using quite different sources of belief about agent play. One example model is based on the game form, and another based on heuristic assumptions of behavior. In both, we assume that the graphical structure representing interactions among players in the game is known. We then introduce and test three approaches to integrate these knowledge sources into a combined model. To evaluate the results, we posit that actual play is generated by agents who start to play heuristically, then update their behavior over repeated interactions through a reinforcement learning (RL) process. Thus, the task we set up is to predict the outcome of a RL regime. More precisely, we seek to compute a reasonable estimation of the joint probability distribution of the agents' play. Intuitively, knowledge about the heuristic starting point is relevant, as is knowledge of strategically stable policies (game-theoretic equilibria), but neither directly captures nor necessarily corresponds to the RL outcome. We find experimentally that in fact the two knowledge sources are complementary, as the combined model outperforms either alone. Our investigation further provides support for one particular combination approach, based on mixing data.

We begin with formal definitions and specifications of the GMM framework in Section 2. In Section 3, we present an example multiagent domain, the Internet industry partnership network, and construct two plausible models based on different knowledge sources. Section 4 details some alternative combination methods. We follow with an empirical study in Section 5, designed to evaluate the performance of the respective models and their combinations. We conclude the paper with some observations on these results.

## 2 GRAPHICAL MULTIAGENT MODELS

Consider a multiagent scenario with $n$ players, where each player $i \in \{1, \ldots, n\}$ chooses an action (or *strategy*) $s_i$, from its strategy domain, $S_i$. The outcome is a joint action, or *strategy profile*, $s$, designating the strategy choice of all players. A GMM $G$ for this scenario is a graphical model, $G = (V, E, S, \pi)$, with vertices $V = \{v_1, \ldots, v_n\}$ corresponding to the agents (we refer to $v_i$ and $i$ interchangeably), and edges $(i, j) \in E$ indicating a local interaction between $i$ and $j$. The graph defines for each agent a neighborhood, $N_i = \{j \mid (i, j) \in E\} \cup \{i\}$, including $i$ and its neighbors $N_{-i} = N_i \backslash \{i\}$. Each neighborhood $i$ is associated with a potential function $\pi_i(s_{N_i}) : \Pi_{j \in N_i} S_j \to \mathbb{R}$. Intuitively a local configuration of strategies with a higher potential is more likely to be part of the global outcome than one with lower potential. As in graphical games, the size of the GMM description is exponential only in the size of local neighborhoods rather than in the total number of players. These local potentials define the joint probability of a global configuration $s$,

$$\Pr(s) = \frac{\Pi_i \pi_i(s_{N_i})}{Z}, \qquad (1)$$

where $Z$ is a normalization term.

One source of potential functions is a description of the *game* played by the $n$ agents. Let $\mathbf{G}$ be a game with agents and strategy sets as defined for the GMM $G$. Let us further assume that $\mathbf{G}$ is a graphical game, such that agent $i$'s payoff depends only on $s_{N_i}$: its strategy and those of its neighbors. Formally, $i$'s payoff is defined by a utility function, $u_i : \Pi_{j \in N_i} S_j \to \mathbb{R}$. For example, Daskalakis and Papadimitriou (2006) defined a binary potential function, associating a high value for configurations where each agent's strategy choice is a best response to its neighbors (i.e., maximizes payoffs given $s_{N_{-i}}$), and a low value for all other configurations.

A natural generalization of this approach would smooth out the binary distinction, assigning intermediate potentials based on the *degree* to which agents deviate from their best response. We may not wish to assume that agents play best responses with certainty, as they may not be perfectly rational, or our attributions of payoff functions may be inexact. For a given payoff model, let $\epsilon_i(s_{N_i})$ denote $i$'s *regret* function, representing the maximum gain $i$ can obtain through unilaterally reconsidering its own strategy $s_i$ given $s_{N_{-i}}$,

$$\epsilon_i(s_{N_i}) = \max_{s'_i \in S_i} u_i(s'_i, s_{N_{-i}}) - u_i(s_{N_i}).$$

Intuitively, we expect that high-regret profiles are less likely to be played (all else equal), since as regret increases agents are more apt to recognize and select the better alternatives. We can capture this intuition in a *regret potential*,

$$\pi_i(s_{N_i}) = e^{-\epsilon_i(s_{N_i})/T_i}, \qquad (2)$$

where $T_i$, the *temperature* parameter, provides a way to calibrate our association between regret and relative likelihood. Greater values of $T_i$ accord more likelihood to agents making less than perfectly rational decisions. For simplicity in notation below, we also define $\lambda_i = \frac{1}{T_i}$, and $\lambda$ the vector of $\lambda_i$.

Let *reG* denote a GMM employing the regret potential function. In the current study, we consider the regret GMM *reG* as one plausible form of predictive model. We also consider models that are not based directly on payoffs in an associated game. In particular, we construct for our particular example a rule-based GMM, *hG*, encoding heuristic assumptions of agents' behavior.

# 3 EXAMPLE: INTERNET INDUSTRY PARTNERSHIPS

We illustrate the GMM framework and motivate the problem of combining knowledge sources through an example multiagent scenario. In the *Internet industry partnership* domain,[1] companies must decide whether to retain ($s = 1$) or upgrade ($s = 2$) their current technology. The payoff functions in $\mathbf{G}$ can be mapped into $G$'s potential functions in several different ways. The payoff for each strategy depends on the choices of other companies to which they are related through some kind of partnership—their neighbors in the interaction graph. For example, the benefits of upgrading may be larger when one's partners also upgrade, since keeping their technologies synchronized enhances compatibility.

## 3.1 GAME DEFINITION

We construct our example scenario using a fragment of the partnership network consisting of 10 representative companies, as depicted in Figure 1. Each node represents a company, which we characterize by three parameters: (1) size class, $z$; (2) sector, $t$: either commerce, infrastructure, or content; and (3) change coefficient, $ch \in [0, 1]$, representing the intrinsic adaptability of the company's technology. The study by Krebs (2002) provides sector and a rough order of size; the change coefficient is assigned by us in an arbitrary manner. Although somewhat contrived, the scenario specification serves our purpose of demonstrating some capabilities of the GMM approach.

The payoff function, $u$, defines the value of retaining or upgrading technology, given the actions of a firm's neighboring companies and the parameters describing these companies. Qualitatively, payoff is increased by agreement with neighbors, where larger neighbors from the same sector are relatively more important. Let us first introduce an intermediate value $w_{ij}$ for each connected pair of companies, reflecting the strength of $i$ and $j$'s partnership:

$$w_{ij}(s_i, s_j) = (z_i + z_j)\left(1 + \frac{y_{ij}}{2^{I_t} + 4^{1-I_t}}\right)^{I_s}, \qquad (3)$$

[1]The example is inspired by the network model of Krebs (2002), cited by Kearns (2002).

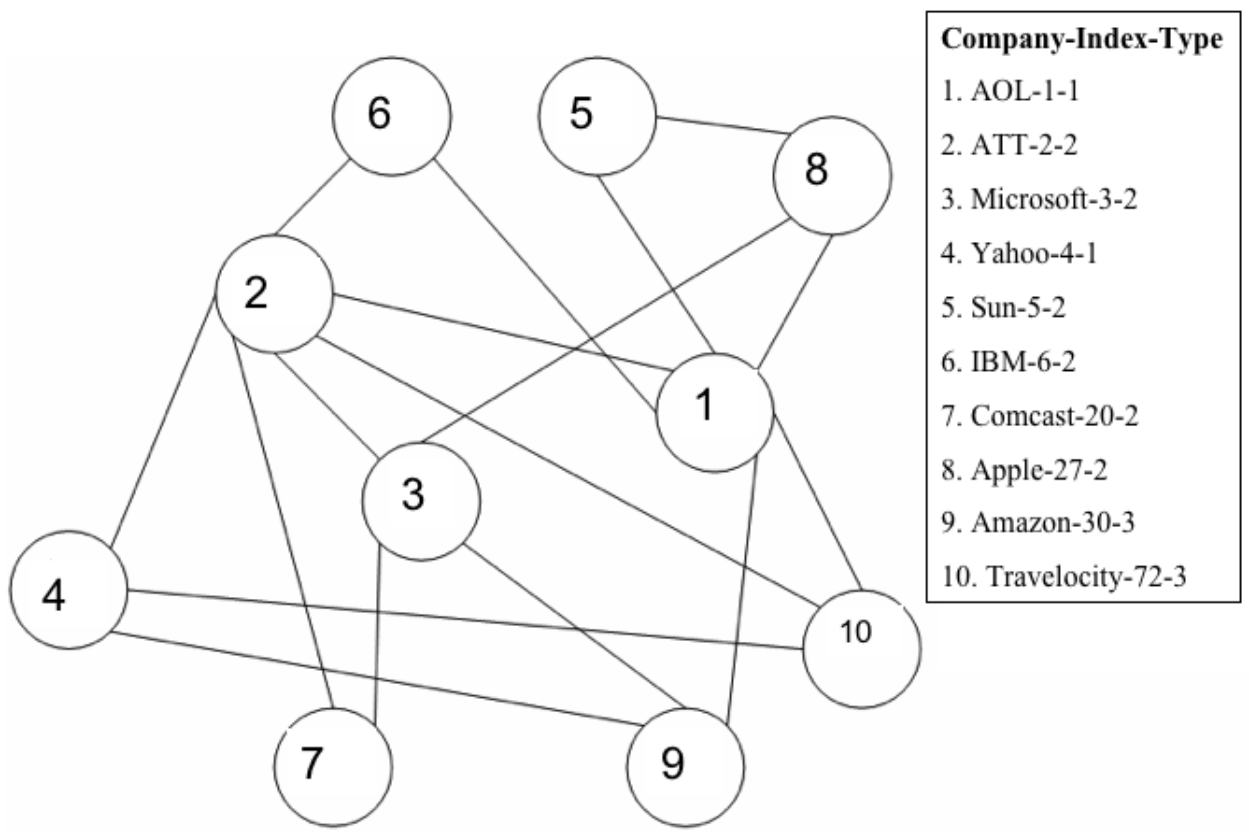

Figure 1: Part of the Internet industry partnership network, from Krebs (2002).

where $y_{ij} \sim U[0, 1]$ is a random variable, and $I_s$ and $I_t$ are indicator functions representing agreement in strategy and technology sector. $I_s = 1$ if $s_i = s_j$, and 0 otherwise; $I_t = 1$ if $t_i = t_j$, and 0 otherwise. The intuition for (3) is that the mutual influence between two firms increases with their size, agreement in action, and sector commonality. We also define function $\phi(ch_i, s_i)$ to compare $i$'s action to its change parameter $ch_i$. Specifically, $\phi$'s value is positive if $s_i$ is upgrade (retain) and $ch_i$ is greater (smaller) than 0.5. The interpretation is that a value greater (smaller) than 0.5 for $ch_i$ implies that $i$ is flexible (inflexible) with respect to technology change.

$$\phi(ch_i, s_i) = \begin{cases} \frac{s_i}{2} - ch_i & \text{if } \frac{s_i}{2} - ch_i < 0.5 \\ 0.5 - \frac{s_i}{2} + ch_i & \text{otherwise} \end{cases}$$

Finally, the overall payoff combines the pairwise partnership weights, further adjusted by the company's flexibility in upgrading its technology.

$$u_i(a_{N_i}) = (1 + y_i \phi(ch_i, s_i)) \sum_j w_{ij}(s_i, s_j),$$

with $y_i \sim U[0, 1]$.

## 3.2 GMM CONSTRUCTIONS

Given the payoff function and the game definition above, we can generate a regret potential function (2) in a straightforward manner, parametrized by temperature. This potential function in turn defines a regret GMM, *reG*.

Our heuristic rule-based model, *hM*, in contrast, is specified without direct reference to the payoff function. In this model, each company independently applies a local heuristic to stochastically choose its action. Specifically, agent $i$ changes its technology with probability $pChange(i)$, where

$$pChange(i) = 0.5(1 - 10^{-3})^{|N_i|}(1 - 10^{-3} z_i).$$

The intuition behind this heuristic is that the more partners ($|N_i| - 1$) a company has and the greater its size $z_i$, the less likely it is to change. Given the $pChange$ values, it is straightforward to define a potential function for the GMM $hG$ such that the outcome distribution is the same as generated by applying the rule independently for each company. As a result, $hG$'s potential $\pi_i$ is a function of only $s_i$ instead of $s_{N_i}$.

The two GMMs, $reG$ and $hG$, are based on qualitatively different sources of knowledge. If we believe that agents are essentially rational and aware of their environment, we may expect that the regret GMM $reG$ would predict behavior well. If instead we have evidence that agents choose heuristically based on partnership density and size, we might have greater confidence in predictions based on $hG$, or on some other heuristic models that capture our intuition of agents' behavior. In other situations, we may consider that both models capture factors determining behavior, and view the knowledge sources as complementary.

### 3.3 SIMULATION MODEL

The role of our simulation model is to generate play data from a plausible agent interaction process. In this study, we treat this data as the actual outcome, and use it to evaluate the GMMs above as well as combined models. The simulation is based on the idea that actual agent behavior is produced via repeated interaction through a reinforcement learning (RL) procedure.

In the model, each agent is an independent learner, employing an RL procedure designed for partially observable environments (Jaakkola et al., 1995). The environment for company $i$ comprises $i$ and its partners (i.e., its neighbors $N_{-i}$), and each configuration of partner strategies $s_{N_{-i}}$ is a possible state. The agent seeks to learn a stochastic play policy: $\sigma_i(s_i \mid s_{N_{-i}})$, which denotes the probability of playing action $s_i$ at state $s_{N_{-i}}$. However, the agent does not actually observe $s_{N_{-i}}$ before taking its action. Thus, the action is actually selected based on the policy, for $a = 1, 2$,

$$\Pr_i(s_i = a) = \sum_{s_{N_{-i}}} \sigma_i(s_i = a \mid s_{N_{-i}}) \Pr(s_{N_{-i}}). \quad (4)$$

$\Pr_i(s_{N_{-i}})$ is a stationary probability distribution over states, which, for simplicity, we take to be uniform. The scarcity of companies' knowledge about others' strategies and the network effects motivates our choice of $\Pr_i(s_{N_{-i}})$ instead of a more complicated model of network dynamics.

To learn the policy $\sigma_i$, we apply the RL procedure of Jaakkola et al. (1995).

1. Initialize $\sigma_i$ to $pChange(i)$ (i.e., the heuristic policy from $hM$).
2. Generate a play $s$ using (4). Observe the resulting local state $s_{N_i}$ and receive as reward the payoff $u_i(s_{N_i})$. Update $Q_i(s_i, s_{N_{-i}})$, the average reward for taking action $s_i$ in the associated local state.
3. Choose $\sigma_i^*(s_i \mid s_{N_{-i}})$ to maximize $Q_i(s_i, s_{N_{-i}})$. Adjust the current policy $\sigma_i$ in the direction of $\sigma_i^*$: $\sigma_i \leftarrow \sigma_i(1-\gamma) + \sigma_i^* \gamma$, where $\gamma$ is the learning rate.
4. Repeat steps 2 and 3 until convergence.

For our experiments, we used $\gamma = 0.2$ and iterated steps 2 and 3 above 40 times, the point after which few changes occurred in the learned RL policy $\sigma$. We denote the simulated model at the end of the RL procedure as $simM$.

Note that the RL process starts with the local heuristic rule-based policy, but is updated based on payoff experience. Thus we expect that both the heuristic rule-based model and the rationalistic regret-based model may offer value for predicting the outcomes of $simM$.

## 4 METHODS FOR COMBINING KNOWLEDGE SOURCES

Given two complementary sources of knowledge, how can we integrate them into a single GMM? We formulate the problem for the case that one knowledge source is expressed explicitly as a GMM, $G_1$, and the other in the form of some data $D = \{s^1, \ldots, s^m\}$ of joint plays related to the multiagent scenario.[2] Note that $D$ may not reflect the actual distribution of play accurately, for example because it is small in size or because it was observed in a different multiagent setting. In this section we answer these questions abstractly and in the next section show how we can do this for our specific $reG$ model and data $D$ derived from $hM$.

### 4.1 DIRECT UPDATE

The *direct update* method combines the two sources of knowledge $G_1$ and $D$ into a new GMM, *directG*, derived by adjusting the $\lambda_{G_1}$ parameters of $G_1$ to maximize the predictive performance w.r.t. the data $D$.

[2]Since we can generate such a data set from a GMM, or induce a GMM from data, the combination methods can be applied to more general settings where knowledge sources come in either form.

We measure predictive performance using the logarithmic scoring rule (Gneiting and Raftery, 2007): $\text{Score}(G \mid D) = \sum_{k=1}^{|D|} \log \Pr_G(s^k)$, which assesses the log-likelihood of the data. We take as our problem to tune the GMM's parameters $\lambda$ in order to maximize this score (Kappen and Rodríguez, 1997),

$$\text{Score}(G_1 \mid D) = \sum_{k=1}^{|D|} \log \Pr_{G_1}(s^k) = L(D \mid \lambda = \lambda_{G_1}).$$

We employ the gradient ascent method to maximize data likelihood, which entails computing the gradient of $L$ w.r.t $\lambda$:

$$\begin{aligned} \nabla\lambda &= \frac{\partial L(D|\lambda)}{\partial\lambda} \\ &= \frac{\sum_k^{|D|} \partial \log e^{-\sum_i \lambda_i \epsilon_i(s^k_{N_i})}}{\partial\lambda} - |D| \frac{\partial \log Z}{\partial\lambda} \end{aligned} \quad (5)$$

and adjusting $\lambda = \lambda + \alpha\nabla\lambda$, where $\alpha$ is the learning rate, until the gradient is below some threshold.

A major problem in graphical-model parameter learning is the intractability of calculating $\log Z$ in (5). It entails iterating over all possible outcomes of the game, which is exponential in the number of players $N$, rendering exact inference and learning in undirected graphical models intractable. Since our priority is to accurately evaluate knowledge combination methods for GMMs, our current implementation of the inference and learning algorithms does not employ any approximation. Our pilot study of generalized belief propagation approximations in GMMs (Yedidia et al., 2001) has indeed yielded positive results, and will be incorporated in future reports.

### 4.2 OPINION POOL

Unlike direct update, which depends on the availability of both play-outcome data and the potential function's parameterized form, the next two methods, *opinion pool* and *mixing data*, push the knowledge combination problem towards potentially greater independence from the input knowledge sources' forms.

The *opinion pool* method starts by first using half the given data $D$ to learn a GMM $G_2$ by adjusting its parameters to maximize the likelihood of the data, as in the direct update method above. The combined model $OPG$ is then an aggregation of $G_1$ and $G_2$ into a single probability distribution:

$$\Pr_{OPG}(s) = f(\Pr_{G_1}(s), \Pr_{G_2}(s)).$$

Note that the above equation does not involve defining a separate pooled potential function for each player in the combined model. The rationale is that potentials are not normalized like the joint probability, and thus, their absolute values contain little meaning when taken out of the context of their corresponding models.

We adopt for our aggregation function the weighted geometric mean, called the *logarithmic opinion pool* (logOP).

$$\Pr_{OPG}(s) = \frac{\Pr_{G_1}(s)^w \Pr_{G_2}(s)^{1-w}}{Z}.$$

The logOP is the only pool known to preserve independence structure in graphical models (Pennock and Wellman, 2005), which is an important property in our context. The weight parameter $w$ can be set by fiat, or tuned with data. In our experiments described below, we employ the other half of input game-play data $D$, denoted $\bar{D}$, and set $w$ by maximizing the objective function:

$$L(\bar{D} \mid w) = \sum_{k}^{|\bar{D}|} \log \Pr_{OPG}(s^k). \quad (6)$$

In brief, given the two components $G_1$ and $G_2$, the opinion pool method first initializes $w = 0.5$. It then repeats computing the gradient $\nabla w$ of the log-likelihood w.r.t $w$ by differentiating (6), and updating $w = w + \beta\nabla w$, where $\beta$ is the learning rate, until the gradient is acceptably small.

### 4.3 MIXING DATA

The *mixing data* method samples joint plays from the given GMM $G_1$ to generate a new data set $D_1$. We combine $D_1$ and $D$ into one data set $mD$ by sampling from the two sources equally, though one could easily weight one source more than another by adjusting the sampling ratio. We then induce a new GMM for a given parametrized form by tuning the parameter $\lambda$, as detailed in the direct update approach (Section 4.1), to maximize the likelihood of the new data $mD$.

Below is the outline of the mixing data method, which produces the combined GMM $mixG$. Note that we leave out step 4 in our implementation.

1. Generate a sample of play outcomes $D_1$ from $G_1$.
2. Build the mixed data set $mD$ from $D_1$ and $D$ with sampling ratio $\omega = 0.5$.
3. Initialize $mixG$ with some $\lambda_{mixG}$. Update $\lambda_{mixG}$ as in direct update using the data $mD$.
4. (optional) Tune $\omega$ in the direction determined by the gradient-descent method to maximize $mixG$'s performance on a small held-out part of the testing data. Repeat steps 3 and 4 until the gradient is below some threshold.

## 5 EMPIRICAL STUDY

We evaluate our combination approaches experimentally using our simplified version of the Internet industry partnership domain. We concentrate mainly on Example 1, the scenario depicted in Figure 1. In the first experiment, we also examine Example 2, which employs a smaller graph including only the top four companies.

### 5.1 EXPERIMENT SETTINGS

In our experiments we use the following components defined above: (i) the regret GMM $reG$ with the temperature parameters generated uniformly randomly (except in one case explicitly specified below), (ii) a data set $D$ of joint plays generated by using the heuristic rule-based model $hM$, (iii) the heuristic model $hM$ and associated GMM $hG$, and (iv) a testing data set $D^*$ derived from the RL-based model $simM$. We compare our different combination approaches based on their ability to predict the test data set as measure by the score function $\text{Score}(G \mid D^*)$. In particular, we compare the performance of a model that combines knowledge sources $combinedG$ relative to the performance of a baseline model $baseG$ using a ratio of scores, $R = \frac{\text{Score}(baseG|D^*)}{\text{Score}(combinedG|D^*)}$.

The inverted order of $baseG$ and $combinedG$ is due to Score's negativity, and thus, any $R > 1$ indicates $combinedG$'s improvement over $baseG$.

We experiment with several environment settings. For each setting, we conduct 20 trials, each of which involves a training data $D$ set of 500 plays from $hM$ and a testing data set $D^*$ of 500 plays from $simM$.

### 5.2 RESULTS AND ANALYSIS

First, in Figure 2 and Figure 3 we present an overview of our combination methods' effectiveness. For both figures, $reG$ and $D$ are used to derive the model $directG$ using the direct update combination method, the model $OPG$ using the opinion pool method, and the model $mixG$ using the mixed data method.

Figure 2 displays predictive performance across the two examples (1 and 2) and the two baseline models ($reG$ and $hG$). Mixing data is consistently best of the three combination methods, and direct update performs relatively better than opinion pool. All three methods yield better results than individual input models, suggesting that combining two knowledge sources is beneficial regardless of which of the proposed methods is adopted.

Figure 3 shows the performance of various models compared to a model derived from the same gold-standard source $simM$ as our test data $D^*$. We sample a separate data set $D'$ from $simM$ and employ it in learning a GMM $simG$ of the parametrized regret form (2), using maximum likelihood to adjust the temperature parameter. The results reveal that our combined models, especially $mixG$, closely match $simG$ in terms of predictive performance.

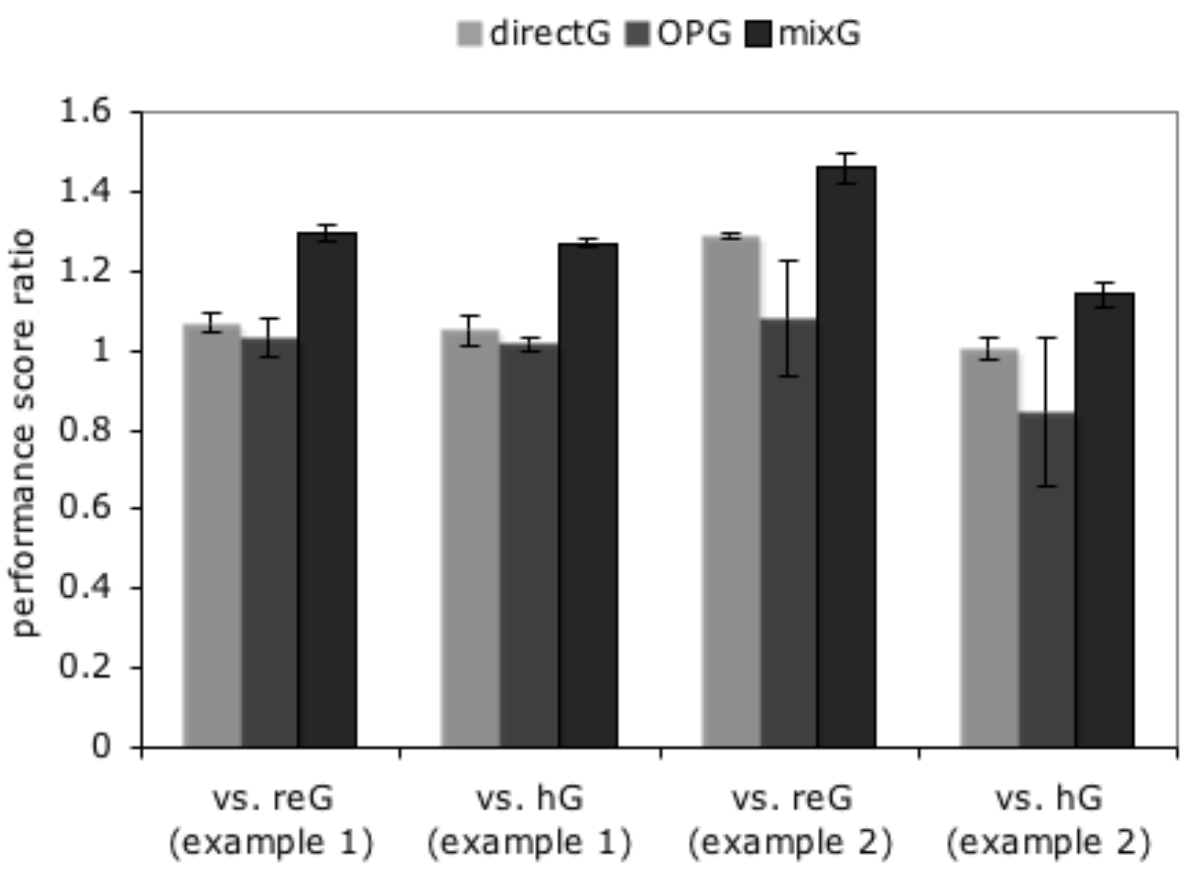

Figure 2: Combination methods' performance across different examples and baselines.

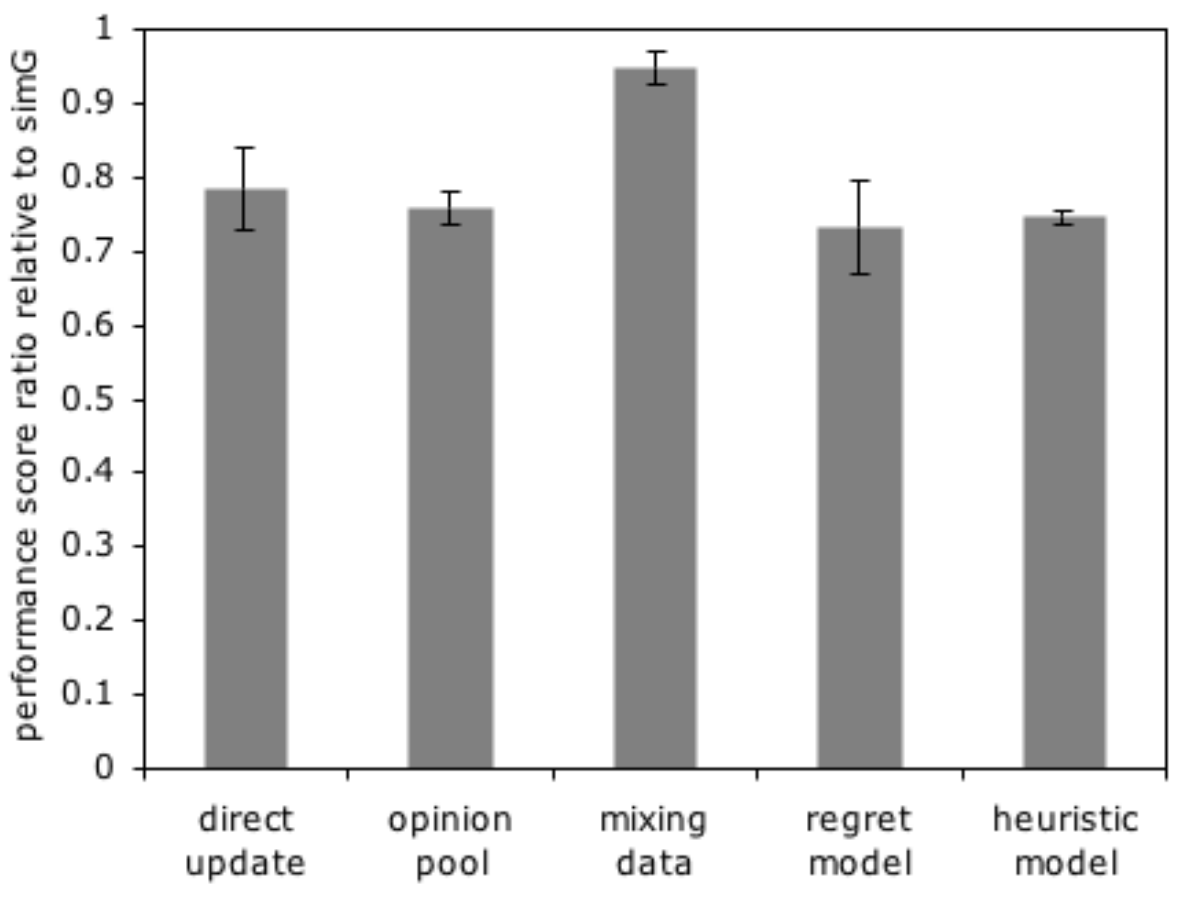

Figure 3: Combination and input models versus the underlying model.

Next, we study the effect of varying the quality of the two input sources. Figure 4 shows the effect of varying the amount of joint-play data available in $D$. Specifically, we make a fraction $\rho|D|$ of play observations, $\rho \in [0, 1]$, available to the three combination methods. From Figure 4, we observe that as long as $\rho > 0.1$, performance remains fairly stable. In our experiments, this corresponds to a threshold data set size of approximately $500 \times 0.1 = 50$. When the amount of data goes below this threshold, the combined model may very well become inferior to the models $reG$ and $hG$.

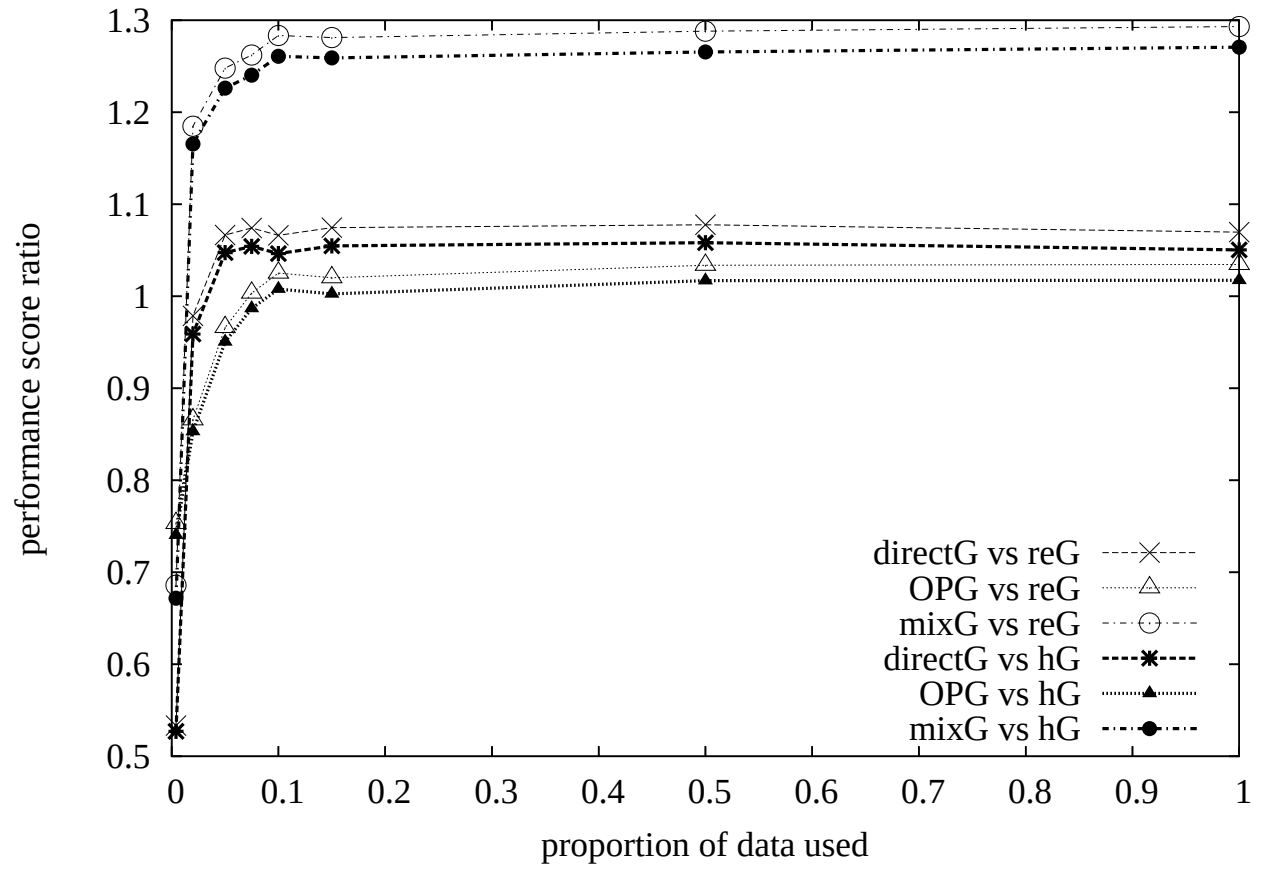

Figure 4: Combination methods' performance versus data availability portion $\rho$.

Figure 5 shows the effect of varying the quality of the GMM provided as input to the combination methods. To modulate the accuracy of $reG$, we introduce a parameter $\delta$ controlling the relation of its temperature parameters to those of $simG$. Specifically, we set $\lambda_{reG}$ to $(1+\delta)\lambda_{simG}$. The results of the third experiment are depicted in Figure 5. When compared with the unchanged heuristic model $hG$, the combination models show a slight decrease in their relative performance with $\delta$, which reflects the effect of $reG$'s inaccuracy on the combination methods. When the baseline is $reG$, in contrast, the degradation of the combined model is dominated by the effect of compromising the baseline $reG$. In other words, combining knowledge sources effectively compensates for degrading one of them.

In these experiments, $OPG$'s poor performance relative to that of $directG$ and $mixG$ may be due in part to its reliance on only a single parameter, $w$, compared to the vector $\lambda$ available to the other methods. $mixG$'s overall superiority is likely a result of its directly sampling from $reG$, which employs information contained in $reG$ more effectively than $directG$, where $reG$ only matters at the initialization stage.

Figure 6 presents the results of an experiment designed to strengthen our claims about the benefits of integrating knowledge sources in a single model. We examine the combined models' performance in environments where the simulation mode $simM$ is not the product of RL that starts with the input model $hM$. In particular, we define a different heuristic model $hM'$, such that $pChange_{hM'}(i) = 0.05$ for all $i$. Let $E$ be the input data set generated from $hM'$. Based on $hM'$, we subsequently build the simulation model $simM'$ and its corresponding test data $E^*$. Given these data sets and models, we can evaluate the combined models across drastically different starting points, by comparing their performances when different input sources, $D$ and $E$, are provided, on the same testing data (either $D^*$ or $E^*$). First, we compare the performance of the heuristic models employed in generating input data, $hG_{(D)}$ (induced from $hM$) and $hG_{(E)}$ (induced from $hM'$): $hG_{(D)}$ performs 60% better than $hG_{(E)}$ when tested on $D^*$, whereas $hG_{(E)}$ outperforms $hG_{(D)}$ by 54% on $E^*$. This assessment affirms that the two different input data sets, $D$ and $E$, are indeed differentiable in terms of the behavior models they represent.

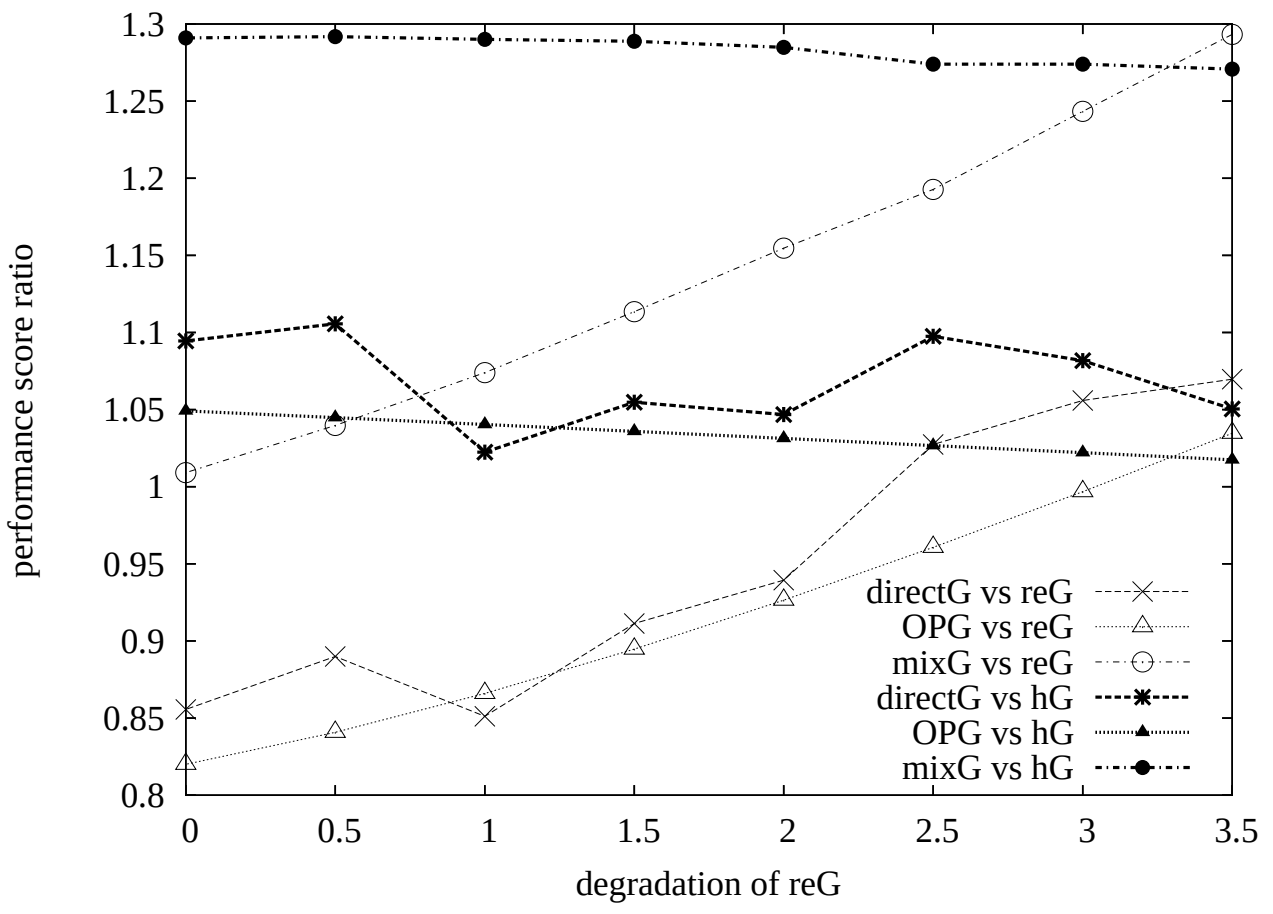

Figure 5: Combination models' performance versus input regret model's inaccuracy as controlled by $\delta$.

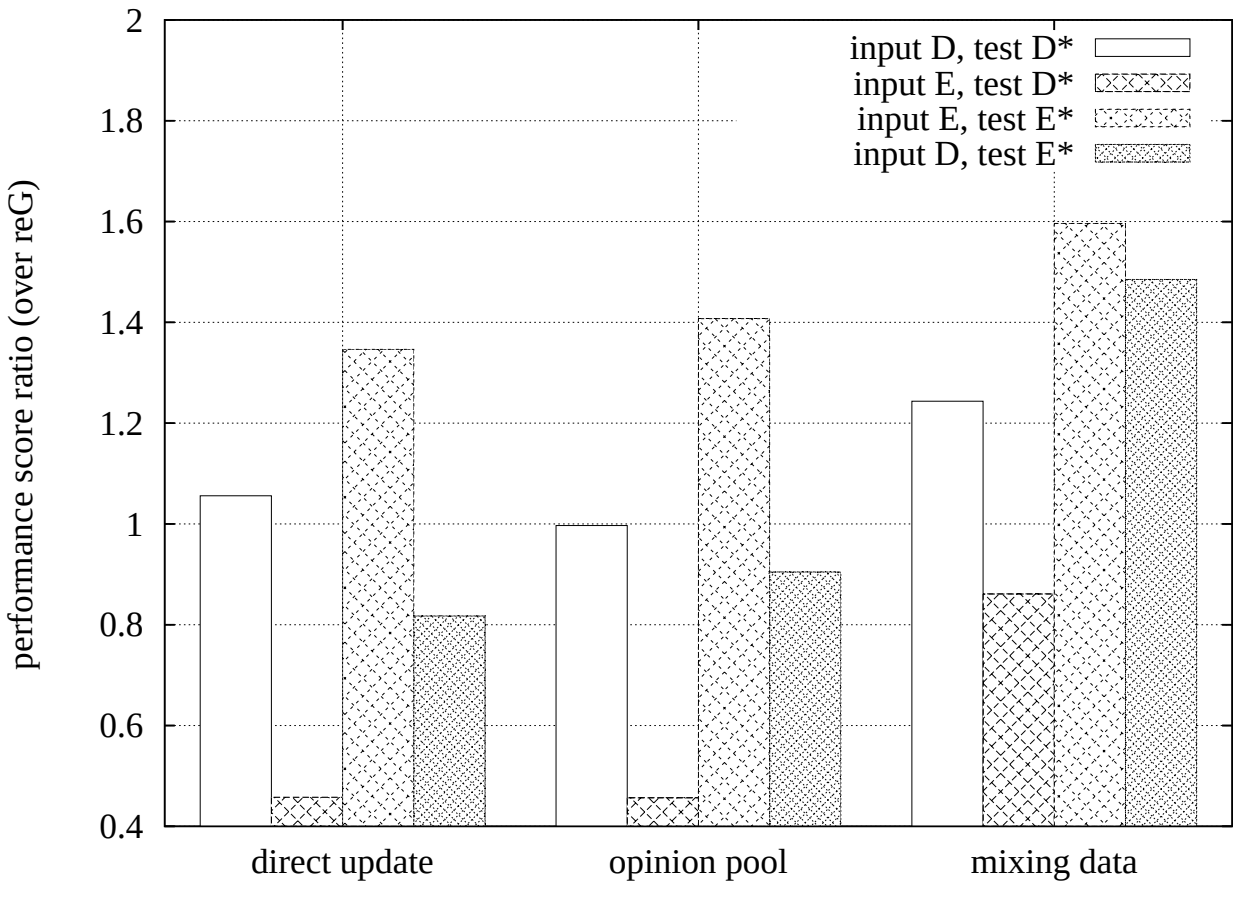

Figure 6: Combination methods in different input and test settings.

The results presented in Figure 6 indicate that better performance is achieved in cases where the test and input environments coincide (the first and third measures) compared to those in which they are unrelated (second and fourth). This observation confirms that combined models whose input data contains some information about the underlying behavior model outperform those extracting irrelevant information from its inputs. The gaps in $mixG$'s performance between scenarios where test cases are derived from the same input (first and third) and from an unrelated model (second and forth) are relatively smaller than those in the other methods. This phenomenon is likely a result of $mixG$'s more effective usage of both knowledge sources, which limits the impact of irrelevant input data, and possibly contributes to the good performance of $mixG_{(D)}$, which is tuned to fit $D$, when tested on $E^*$.

## 6 CONCLUSIONS

GMMs provide a flexible representation framework for graphically structured multiagent scenarios, supporting the specification of probability distributions based on game-theoretic models as well as heuristic or other qualitatively different characterizations of agent behavior. We explored the possibility of exploiting this flexibility by employing multiple knowledge sources for prediction, as demonstrated for the task of predicting the outcome of a reinforcement learning process.

Our basic finding is that combining two knowledge sources in this scenario does improve predictive power over either input source alone. We have also identified the most effective combination method among those tried—mixing data—and the existence of a threshold for data availability that can help boost efficiency. Furthermore, we have found that our knowledge combination approaches, especially mixing data, can effectively match the performance of modeling the reinforcement learning process directly.

This study is a first step in what we expect to be an extended effort to develop the GMM framework for supporting reasoning about strategic situations. One important issue to address is computational feasibility; although the graphical representation facilitates scalability in the number of agents, accurate approximation techniques are nevertheless essential to support practical applications with large models.

Knowledge about multiagent behavior may come from sources other than play history and regret functions, and so another logical research direction is to develop canonical models for capturing and combining such sources. Learning may be extended to not only the model's parameters, but also the structure of potential functions and the graph topology itself. Such extensions present more complicated problems for combining models derived from different knowledge sources. Finally, we might look to dynamic Bayesian network concepts to extend the GMM framework to add a time dimension, and thus enable modeling sequential and interactive multiagent environments.